\documentclass{article}

\usepackage{arxiv}

\usepackage[utf8]{inputenc} 
\usepackage[T1]{fontenc}    
\usepackage{hyperref}       
\usepackage{url}            
\usepackage{booktabs}       
\usepackage{amsfonts}       
\usepackage{nicefrac}       
\usepackage{microtype}      
\usepackage{lipsum}
\usepackage{graphicx}
\usepackage{subfigure}
\usepackage[ruled,vlined]{algorithm2e}
\usepackage{threeparttable}  
\usepackage{multirow}

\title{Dynamic Pricing on E-commerce Platform with Deep Reinforcement Learning: A Field Experiment}

\author{
  Jiaxi Liu \\
  Alibaba Supply Chain Platform\\
  \texttt{galiliu.ljx@alibaba-inc.com} \\
   \And
 Yidong Zhang\thanks{Correspondence: 969 West Wen Yi Road, Yu Hang District, Hangzhou, China, 311121} \\
  Alibaba Supply Chain Platform\\
  \texttt{tanfu.zyd@alibaba-inc.com} \\
  \AND
  Xiaoqing Wang\\
  Alibaba Supply Chain Platform\\
  \texttt{robin.wxq@alibaba-inc.com} \\
  \And
  Yuming Deng\\
  Alibaba Supply Chain Platform\\
  \texttt{yuming.dym@alibaba-inc.com} \\
  \And
  Xingyu Wu\\
  Alibaba Supply Chain Platform\\
  \texttt{zhuyang.wxy@alibaba-inc.com} \\
}

\begin{document}
\maketitle

\begin{abstract}
In this paper we present an end-to-end framework for addressing the problem of dynamic pricing (DP) on E-commerce platform using methods based on deep reinforcement learning (DRL). By using four groups of different business data to represent the states of each time period, we model the dynamic pricing problem as a Markov Decision Process (MDP). Compared with the state-of-the-art DRL-based dynamic pricing algorithms, our approaches make the following three contributions. First, we extend the discrete set problem to the continuous price set. Second, instead of using revenue as the reward function directly, we define a new function named difference of revenue conversion rates (DRCR). Third, the cold-start problem of MDP is tackled by pre-training and evaluation using some carefully chosen historical sales data. Our approaches are evaluated by both offline evaluation method using real dataset of Alibaba Inc., and online field experiment starting from July 2018 with thousands of items, lasting for months on Tmall.com. To our knowledge, there is no other DP field experiment using DRL before. Field experiment results suggest that DRCR is a more appropriate reward function than revenue, which is widely used by current literature. Also, continuous price sets have better performance than discrete sets and our approaches significantly outperformed the manual pricing by operation experts.
\end{abstract}

\keywords{Reinforcement learning \and Dynamic pricing \and E-commerce \and Revenue management \and Field experiment}

Dynamic pricing, to adjust prices according to inventories left and demand response observed, has drawn great attentions during the past decades since the deregulation of the airline industry in the 1970s. \cite{weatherford1992taxonomy} and \cite{talluri2006theory} gave overviews of the research that has been done in the field of perishable-asset revenue management, which is a field that combines the areas of yield management, overbooking, and pricing.

During the recent development of business, many industries have become more active in revenue management. Ride-sharing platforms like Uber has implemented dynamic pricing strategy, known as 'surge' pricing and \cite{chen2016dynamic} showed that it has significant impact on motivations for more driving times. Retailers like Zara have implemented systematic dynamic markdown pricing strategy \cite{caro2012clearance}. Kroger is now testing electronic price tag at one store in Kentucky (\cite{nicas2015now}). 

Online retailers have a stronger desire for dynamic pricing strategies due to the requirement of more complex operations. For example, Amazon.com sells 356 million products (562 million now). Walmart.com sells 4.2 million products according to a 2017 estimate\footnote{https://www.scrapehero.com/how-many-products-does-walmart-com-sell-vs-amazon-com}. Taobao.com, the biggest E-commerce platform in China, sells billions of products at present. Operation specialists have to set prices for these items periodically to remain competitive while maximize revenue, which will be mission impossible when the number of items goes this high. As a result, Amazon has implemented automatic pricing systems and it is reported that Amazon.com can change prices every 15 minutes\footnote{https://www.whitehouse.gov/sites/default/files/docs/Big\_Data\_Report\_Nonembargo\_v2\.pdf}. \cite{chen2016empirical} studied the pricing strategies for Amazon.com. 

In this paper, we proposed a reinforcement learning approach to address the dynamic pricing problem for online retailers. The scenario we consider is how to dynamically price for different products  on Tmall.com, the largest business-to-consumer retailer in China, spun off from Taobao.com. There are many difficulties for pricing on such an E-commerce platform. First, the market environment is impossible to be quantified. Demand for the same product could change dramatically due to unpredictable fluctuation of the daily customer traffic, the change of other products' prices or even comments from the previous buyers. Second, it would lead to non-convergence policies if the reward function is not set properly under such complicated environment. Third, it is not appropriate to apply a learning model online directly, since a slightly inappropriate price online could quickly cause large capital loss. Fourth, unlike recommendation system\cite{davidson2010youtube}, it is impossible to do online A/B testing, because it is illegal to expose different prices at the same time to different customers. That is an obstacle to evaluate the performances of different pricing policies during the field experiment.

To overcome these difficulties, we propose a framework for dynamic pricing with DRL to optimize long-term revenue. This paper has several contributions. First of all, we are the first to apply DRL for both discrete and continuous pricing problem on real-world E-commerce platform, rather than traditional discrete one using Q-learning (\cite{maestre2018reinforcement}) or within simulated environment (\cite{kephart2000dynamic}, \cite{kim2016dynamic}, \cite{maestre2018reinforcement}, \cite{raju2003reinforcement}, \cite{vengerov2007gradient}). To achieve this, the real-world market environment is precisely described with four groups of features we defined. Second, we found that the revenue conversion rate and the difference of the revenue conversion rate are more suitable as the reward function rather than the revenue (\cite{kim2016dynamic}, \cite{maestre2018reinforcement}, \cite{schwind2002dynamic}) due to its concave nature. Third, aiming the cold start problem, we designed pre-training and evaluation procedures within our pricing framework. Fourth, we defined \textbf{simi-products} on E-commerce platform, which we proved in our experiments that they could be used for evaluating different pricing policies. Finally, we carry out large-scale field experiments lasting for months with thousands of SKUs of products priced by our DRL models. Our field experiments indicate the effectiveness of our dynamic pricing framework on E-commerce platform, which is the first work of this kind with such achievement.

The remaining of this paper is organized as follows: The next section lists some related work in dynamic pricing problem. Section \ref{Methodology} introduces approaches we designed for dynamic pricing, where the problem is modeled as a Markov Decision Process model. Both discrete pricing action model and continuous pricing action model are proposed. In section \ref{Experiments}, the results from both offline and online experiments are introduced, which validate our approach of the problem. The conclusions and future work directions are summarized in section \ref{conclusions}. 

\section{Literature review}

Much research has been done in dynamic pricing for decades. We refer to \cite{den2015dynamic} for a comprehensive review for recent developments. 
It combines two research fields: (1) statistical learning, specifically applied to the problem to estimate demand and (2) price optimization. Most of previous research has focused on the cases where a functional relationship between price and demand is assumed to be known to decision makers. \cite{cournot1897researches} is acknowledged to be the first to mathematically describe the price-demand relation of products and solve the mathematical problem to achieve the optimal revenue. However, it assumed that the relationship is static over time which usually does not hold in reality. \cite{10.2307/2300113} assumed that the demand is not only a function of price, but also the time-derivative of price, leading to a dynamic demand function of price over time.
\cite{kamrad2005innovation} introduced a stochastic model to capture demand uncertainty while optimizing the prices. \cite{gallego1994optimal} considered constraints like limited inventories and a finite planning horizon. 

In practice, it is often difficult to describe the demand beforehand. Much recent research focuses on the dynamic pricing with unknown demand function. Some researchers addressed the problem by parametric approaches. \cite{bertsimas2006dynamic} assumed parametric families of demand functions to be learned over time. \cite{farias2010dynamic} proposed an approach to learn from the historical purchase data. \cite{harrison2012bayesian} utilized Bayesian dynamic pricing policies to address demand uncertainties. However, revenue may depart from the optimal due to mis-specifying the demand family. Therefore, much recent research mainly revolves around non-parametric approaches. \cite{besbes2009dynamic}, \cite{besbes2015surprising}, \cite{wang2014close} looked deep inside learning while earning approaches. However, they all assumed the revenue function is strictly concave and differentiable, which could not hold in E-commerce retail industry as shown in section 3.1. 

With the development of computation, reinforcement learning (RL) is introduced to address dynamic problems (\cite{mnih2015human}, \cite{silver2017mastering}). \cite{kephart2000dynamic} demonstrated the possibility of using Q-learning to express anticipated future-discounted profits for possible prices, forming a so called pricebot to adjust prices in response to changing market conditions. \cite{schwind2002dynamic} used Temporal Difference for Information products' dynamic pricing from a yield management view. \cite{raju2003reinforcement} formulated single seller and two sellers dynamic pricing problems and employ different RL algorithms in a simulated context. \cite{kutschinski2003learning} used different types of asynchronous multi-agent RL methods to determine the competitive pricing strategy in the market scenario. \cite{kim2016dynamic} and \cite{vengerov2007gradient} utilized reinforcement learning to optimize prices in the energy market. \cite{maestre2018reinforcement} suggested Q-learning with neural networks approximation to maintain revenue while improving fairness in a simulated environment. All these previous works, however, are carried out in simulations with simplified market settings, where the reward defined with revenue work out well. Meanwhile, DNNs are only used for approximation of discrete prices, which, however, is not the case for real-world market.

\section{Methodology}\label{Methodology}

\begin{figure}[ht]
\centering
\includegraphics[width=\textwidth]{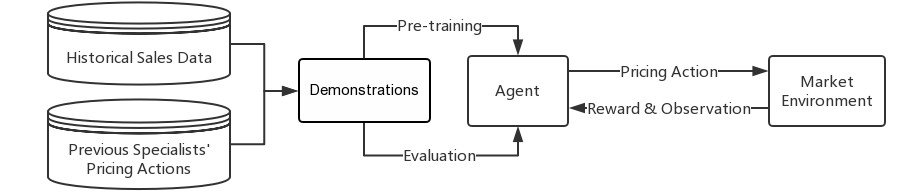}
\caption{Dynamic pricing framework using DRL with demonstrations on E-commerce platform.}
\label{fig:flowchart}
\end{figure}

In our study we first represent the dynamic pricing problem as a Markov Decision Process (MDP). The agent periodically changes prices of the products as its action after observing environment state. The new environment state could then be observed and the reward could also be received. Each pricing episode reaches its end if the product is out of stock. The model is pre-trained by historical sales data and previous specialists' pricing actions, which is also used for offline evaluation. The framework is shown in Figure \ref{fig:flowchart}.

\subsection{Problem formulation}\label{Problem_Formulation}
We mainly consider two kinds of dynamic pricing applications for E-commerce platform, named \textbf{markdown pricing} and \textbf{daily pricing} in the rest of this work. For both markdown pricing and daily pricing, we define $n$ products labeled by $i=1,2,...,n$ individually. The pricing process is formed as Markov Decision Process (MDP) referring works from \cite{kim2016dynamic}, \cite{raju2003reinforcement}, \cite{vengerov2007gradient} etc. Prices are decided to be modified or kept at discrete time steps $t=1,2,...,T$ referring to the model of market in \cite{kutschinski2003learning}, \cite{madhavan2000market} and \cite{o1995market}. The distance between two time steps is defined by a hyper-parameter $d$. At each time step $t$, the pricing agent observes $m$-dimensional state vector $s_{i,t} \in \mathcal{S} \subseteq \mathbb{R}^m$ describing the state of product $i$, and takes an action $a_{i,t}$. Then the agent receives the reward signal $r_{i,t}\in\mathbb{R} $ for that action as well as the observation for the new state $s_{i,t+1}$. These four elements form the transition $(s_{i,t},a_{i,t},r_{i,t},s_{i,t+1})$, which could be simplified as $(s,a,r,s')$. For markdown pricing, the supply is limited, so the pricing process for certain product reaches its end if it is out of stock. For daily pricing application, the supply is regarded as unlimited. 

Intuitively we want a small $d$ to make pricing actions reacting in time or archiving continuous pricing. However, precisely describing the change of the environment may need a certain time period for observing. Changing price rapidly could also break \textit{price image} for the product and even cause credit issue on E-commerce platform. Since our experiments would change the prices online, we set this period carefully after discussions with professional pricing managers. In the rest of this work, pricing period $d$ is settled as one day. Therefore, time step $t$ represents day $t$.

\textbf{State space.} Here in our model, each product $i$ is priced separately, with four different groups of features at time step $t$ to describe the state $s_{i,t}$: price features, sales features, customer traffic features and competitiveness features. Price features contain the actual payment for this product, the discount rate, the coupons, etc. Sales features contain the sales volume, revenue, etc. Customer traffic features contain the time the page of the product $i$ has been viewed, the number of unique visitors viewed the product (UV), the number of buyers for product $i$, etc. The comments and the states of the similar products contribute to competitiveness features. 

\textbf{Action space.} We also define the action space for each product $i$ separately. We use the maximum price $P_{i,max}$ and the minimum price $P_{i,min}$ of the product $i$ during a certain number of periods in the history to define the upper bound and lower bound. It assumes that the pricing framework should not output a price out of this area. The pricing space could be discrete or continuous for different applications. When it is discrete, the pricing space is divided from $P_{i,min}$ to $P_{i,max}$ into $K$ separated areas, as $K$ discrete actions. The price $p_{i,k} \in
[P_{i,min} +\frac{(P_{i,max}-P_{i,min})}{K}\cdot(k-1), P_{i,min} +\frac{(P_{i,max}-P_{i,min})}{K}\cdot k)$
will be regarded as the $k$th ($k \in [1,K]$) pricing action for product $i$. And then each action $a_{i,t}\in\{1,2,...,K \}$ stands for a price range. When it is continuous, action $a_{i,t}\in \mathcal{A} \subseteq \mathbb{R} $ represents an specific price.

\begin{figure} 
\centering
\subfigure[]{
    \begin{minipage}[b]{0.22\textwidth}
    \includegraphics[width=1\textwidth]{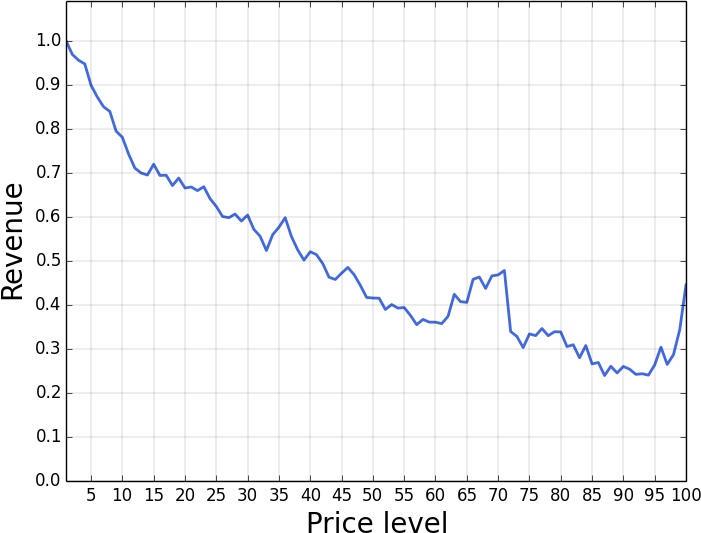}
    \end{minipage}
}
\subfigure[]{
    \begin{minipage}[b]{0.22\textwidth}
    \includegraphics[width=1\textwidth]{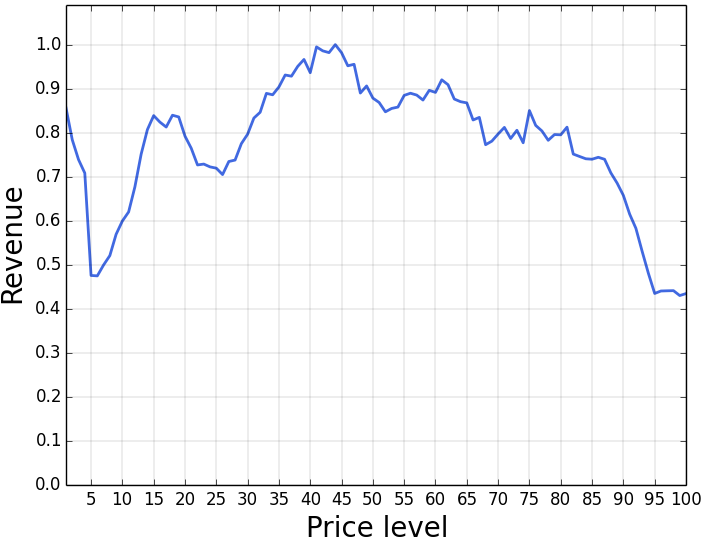}
    \end{minipage}
}
\subfigure[]{
    \begin{minipage}[b]{0.22\textwidth}
    \includegraphics[width=1\textwidth]{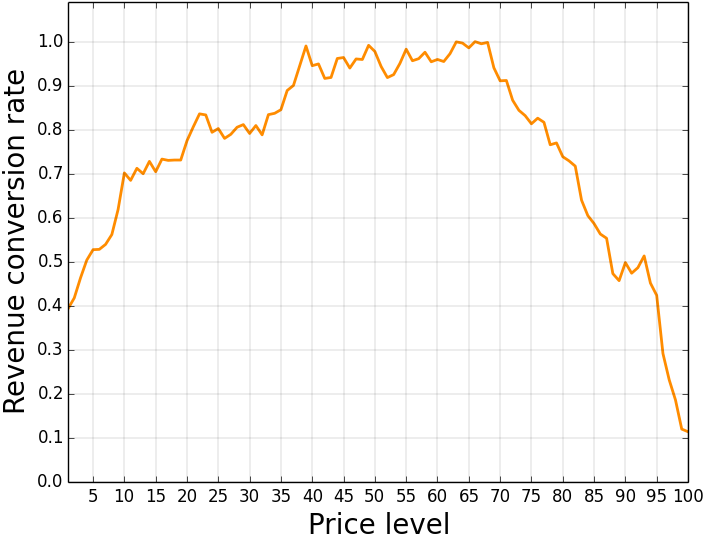}
    \end{minipage}
}
\subfigure[]{
    \begin{minipage}[b]{0.22\textwidth}
    \includegraphics[width=1\textwidth]{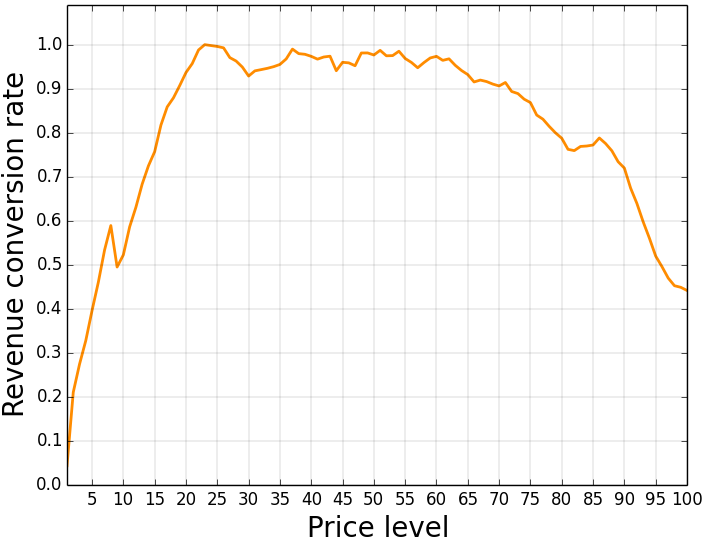}
    \end{minipage}
}
\caption{(a) and (b) show the average of re-scaled revenue for 1400 SKUs of jewelries and 4800 SKUs of candies respectively in different price levels (with the highest revenue re-scaled to 1). Price level 1 stands for the lowest price in three months while price level 100 stands for the highest. (c) and (d) show the average revenue conversion rate for jewelries and candies respectively (with the highest revenue conversion rate re-scaled to 1).  }
\label{fig:revenue}
\end{figure}

\textbf{Reward function.}  We compared revenue with respect to revenue conversion rate to define the immediate reward $r_{i,t}$. The customer traffic could fluctuate drastically and then dominate the total revenue. Therefore, there may not be a clear and explainable relationship between price and revenue like in traditional retail industry. On the other hand, mapping price with revenue could be regarded as a special case of our research assuming the customer traffic is steady. But on E-commerce platform, there are links between prices and revenue conversion rates, $r_{i,t} = revenue_{i,t}/uv_{i,t}$, where $revenue_{i,t}$ and $ uv_{i,t}$ represent the total revenue of product $i$ and the number of unique visitors viewed the product $i$ between time step $t$ and $t-1$ respectively. In some part of this work, we also use profit conversion rate, dividing $profit_{i,t}$ the profit of product $i$ by $uv_{i,t}$, if we have the knowledge of the inventory cost. To prove this idea, we analyze different categories of products about their distribution of revenues and conversion rates on different price levels. Here, we demonstrate the results for over 1400 SKUs of jewelries and over 4800 SKUs of candies selling on Tmall.com in Figure \ref{fig:revenue}. 

\begin{figure}[ht]  
\centering
\subfigure[]{
    \begin{minipage}[b]{0.45\textwidth}
    \includegraphics[width=1\textwidth]{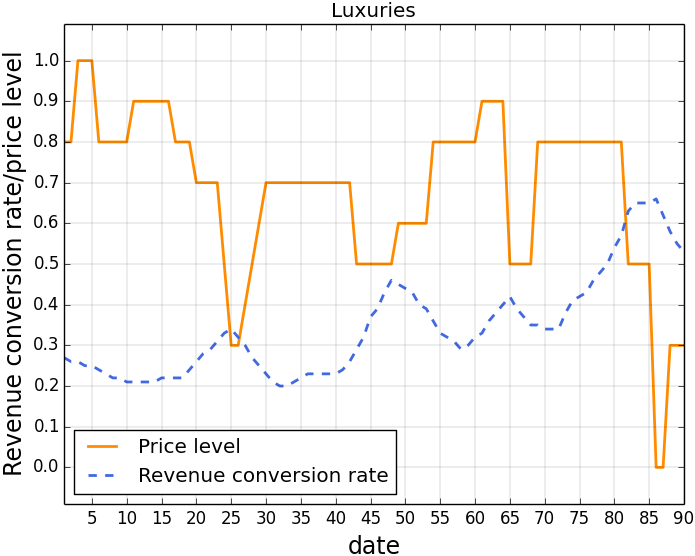}
    \end{minipage}
}
\subfigure[]{
    \begin{minipage}[b]{0.45\textwidth}
    \includegraphics[width=1\textwidth]{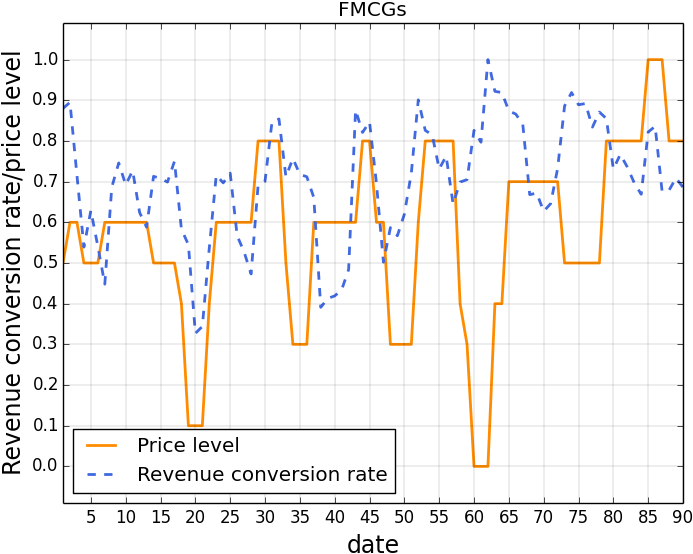}
    \end{minipage}
}
\caption{The average of re-scaled revenue conversion rate and price level for 2000 SKUs of luxuries (in sub-figure (a)) and 4000 SKUs of FMCGs (in sub-figure (b)) respectively through 90 days. Here 0 stands for the lowest price level through 90 days and 1 stands for the highest. The revenue conversion rate is re-scaled by dividing the maximum value. }
\label{fig:price_level}
\end{figure}

As shown in Figure \ref{fig:revenue} that, the revenue conversion rate is more concave than revenue itself. In the field experiments, using revenue conversion rate as reward function works well in markdown pricing application when 1) most of the markdown products are low-sales-volume luxuries, having low but sensitive revenue conversion rates with prices; 2) the revenue conversion rates for these products are relatively steady; 3) there is a very clear and accurate stock determining the length of the pricing process, and the total discounted revenue conversion rate is finite. However, in another pricing application in this work, daily pricing for fast moving customer goods (FMCGs), the supply is adequate and the stock could be regarded as unlimited. A clear end point for each pricing process could hardly be defined. More importantly, the average sales volumes for these FMCGs are much higher than the luxuries and the revenue conversion rates are very unstable. In this case, despite the relationship between prices and revenue conversion rates may not keep steady. Another investigation for two different kinds of products about the trends of their price level with revenue conversion rate for 90 days (shown in Figure \ref{fig:price_level}) reveals this phenomena.

We could see from Figure \ref{fig:price_level} (a) that, for these luxuries, when the price drops, the revenue conversion rate goes up, especially around day 25, 45, 65 and 85. However, for FMCGs in (b) there is no such relationship, but the revenue conversion rate fluctuates with an individual frequency. Within this period, the correlation coefficients between price levels and revenue conversion rates are -0.57 and 0.15 for luxuries and FMCGs respectively. Therefore, if the revenue conversion rate is used as reward function for these FMCGs, the convergence of the model could not be guaranteed. Comparing two phenomenon in Figure \ref{fig:revenue} and Figure \ref{fig:price_level}, we define a different reward function, using the difference of the revenue conversion rates (DRCR) in Eq.(\ref{con:reward_function}):
\begin{equation}
r_{i,t} = \frac{revenue_{i,t}}{uv_{i,t}} - \frac{revenue_{i,t-\tau}}{uv_{i,t-\tau}}, \label{con:reward_function}
\end{equation}
where $\tau$ represents the length of the time to compare the revenue conversion rates. The idea behind this definition is that, we hope to give the agent an positive signal, if it relatively raise the revenue conversion rate with its pricing action.  Our experiments indicate that, this definition of reward function solves the convergence problem in FMCGs daily sales, while could also work out fine in markdown pricing.

\subsection{Discrete pricing action models}
To solve the dynamic pricing MDP we defined above, we first use Q-learning (\cite{watkins1989learning}) to find the optimal pricing policy. Q-learning is a value iteration method to compute the optimal policy. It starts with randomly initialed $Q$ value and recursively iterates using the transitions $t=(s,a,r,s')$ to get the optimal $Q^*$ as well as the optimal policy: 
\begin{equation}
Q_{t+1}(s,a) \leftarrow(1-\alpha)\cdot Q_t(s,a)+\alpha \cdot[ r +\gamma \cdot \max_{a'} Q_t(s',a') ],\label{con:qiteration}
\end{equation}
where $\alpha \in (0,1]$ is the learning rate and $\gamma$ is the discount factor. Due to the high dimension of the state space, we use a deep network to map the Q-values from the state space, which follows the idea of deep Q-networks (DQN, \cite{mnih2015human}). To update the action value network, a one-step off-policy evaluation is used to minimize the loss function:
\begin{equation}
L(\theta) = \mathbb{E}_{(s,a,r,s')\sim D}[r +\gamma \cdot \max_{a'} Q(s',a'|\theta') - Q(s,a|\theta)]^2, \label{con:lossfunction}
\end{equation}
where $D$ is a distribution over transitions contained in a replay buffer working, $\theta$ are the parameters of the Q-network and $\theta'$ are the network parameters used to compute the target. The target network parameter $\theta'$ are only updated with the Q-network parameters every C steps.

\subsection{Continuous pricing action model}
Pricing on discrete action space encounters an obvious conflict setting the number of discrete actions, i.e. the hyper-parameter $K$ in this work. If $K$ is too small, different prices in a large pricing area will be regarded as the same price action for the agent. In return, the discrete output action would also be a large pricing area, making the policy imprecise. On the other hand, if $K$ is too large, a lot of actions will not be explored in the history and the exploration in the future could also be inefficient and expensive. Therefore, we consider to build up model pricing on continuous space to output an exact price instead of a pricing area. We apply the actor-critic algorithm (\cite{witten1977adaptive}), which combines value-iteration methods and policy-iteration methods and have been proposed and performed well on different problems(\cite{mnih2016asynchronous}, \cite{vamvoudakis2010online}). Specifically, we apply deep deterministic policy gradient (DDPG, \cite{lillicrap2015continuous}) as our actor-critic method. The actor part of this model maintains a policy-network $\pi(a|s;\theta^\mu)$, taking the environment state as its input and output continuous actions $a=\mu_\theta (s)$. And the critic part takes both the state and action as input and estimates the action value function $Q(s,a|\theta^Q)$. $\theta^\mu$ and $\theta^Q$ are the network parameters. So, the loss function would be:
\begin{equation}
L(\theta) = \mathbb{E}_{(s,a,r,s')\sim D}[r +\gamma \cdot Q(s',\mu(s'|\theta^{\mu'})|\theta^{Q'}) - Q(s,a|\theta^Q)]^2 \label{con:qloss}
\end{equation}
In this work, we also apply the experience replay and separate target network techniques, so $\theta^{\mu'}$ and $\theta^{Q'}$ are the target-network parameters for actor and critic respectively. And it takes the gradient of Q-value to update the policy-network:
\begin{equation}
\nabla_{\theta^\mu} \mu \approx \mathbb{E}_{\mu'}[\nabla_a Q(s,a|\theta^Q)|_{a=\mu(s)}\nabla_{\theta^\mu}\mu(s|\theta^\mu)].
\end{equation}
The idea is to adjust the parameters $\theta^\mu$ of the policy-network in the direction of the performance gradient, underlying these algorithms is the policy gradient theorem (\cite{sutton2000policy}).
\subsection{Pre-training}
If we directly apply reinforcement learning algorithms to E-commerce dynamic pricing,  the cold start problem occurs, which starts with very poor performance and may cause capital loss. In some other areas like robotics \cite{levine2016end} and games \cite{silver2016mastering}, there may be accurate simulators, within which the agent could learn policy. However, there is no such a simulator for dynamic pricing problem. Instead, we have enough data of the environment and the pricing decisions made by some previous controllers. These controllers could be some specialists or some rules, and some of their pricing decisions may be reasonable. The records of their decision facing the environment could be regarded as demonstrations, which has been proved to be effective for pre-training the agent from \cite{sendonaris2017learning}. 

As mentioned before, the pricing actions were taken periodically. Thus, the state of the environment as well as the rewards could be represented by the data collected within these periods between actions. Therefore, we form the demonstration in tuples $<s_t, a_t, r_t, s_{t+1}>$ and use them for pre-training. Specifically, we refer the ideas of Deep Q-learning from Demonstration (DQfD) \cite{sendonaris2017learning} as our pre-training method for DQN and Deep Deterministic Policy Gradient from Demonstration (DDPGfD) \cite{vecerik2017leveraging} for DDPG.

\subsection{Offline evaluation}\label{para:evaluation_methodology}

As we discussed above, we need to evaluate the model during pre-training before pricing online. We will introduce the methodology for offline evaluation in this part, while the methods for online evaluation will be discussed in detail in section \ref{para:online_experiments}. We first use the latest $T$ periods of records to form tuples $<s_t, a_t, r_t, s_{t+1}>, t\in[1,T]$. And then, we divide these tuples into two parts: the first $D$ tuples, $D<T$, will be used for pre-training, where $t\in[1,D]$. And for $D<t<T$, the tuples will be used for evaluation. The idea is that we sum the reward $r_t$ only if the action $a_t$ is close to the output of the policy $a_t-\epsilon<\pi(s_t)<a_t+\epsilon$. The detail of the evaluate algorithm is sketched in Algorithm \ref{alg:policy_evaluation}.
\begin{algorithm}[htb] 
  \SetKwInOut{Input}{input}
  \SetKwInOut{Output}{output}
  \SetKwRepeat{Repeat}{repeat}{until}
  \caption{ Policy evaluation with demonstrations.}  
  \label{alg:Framwork}  
    \Input{$T>0$: number of demonstration tuples for evaluation.
      $\pi$: the policy to evaluate;}  
    \Output{$R_\pi$: average reward form policy $\pi$;}
    \Begin{ 
        $R \leftarrow 0$, \space $N \leftarrow 0$\;
    \For {step $t \in \{1,2,...,T\}$}{
      \Repeat{$a_t-\epsilon<\pi(s_t)<a_t+\epsilon$}{Get next tuple $<s_t,a_t,r_t,s_{t+1}>$\;}
      $R \leftarrow R + r_t$, \space $N \leftarrow N + 1$\;
    } 
    \If{$N>0$}{ $R_\pi = R/N$}
    \Else{  $R_\pi=0$}
    }
  \label{alg:policy_evaluation}
\end{algorithm}  
\section{Experimental results} \label{Experiments}

In this part, we first introduce the offline experiments, using historical data from Tmall.com. Then we introduce the field experiments. During these field experiments, we changed the selling prices for products on Tmall.com in a markdown scenario and a daily basis starting from July 2018.

\subsection{Offline experiments}\label{para:offline_experiments}

\begin{figure}[ht]  
\centering
\subfigure[]{
    \begin{minipage}[b]{0.3\textwidth}
    \includegraphics[width=1\textwidth]{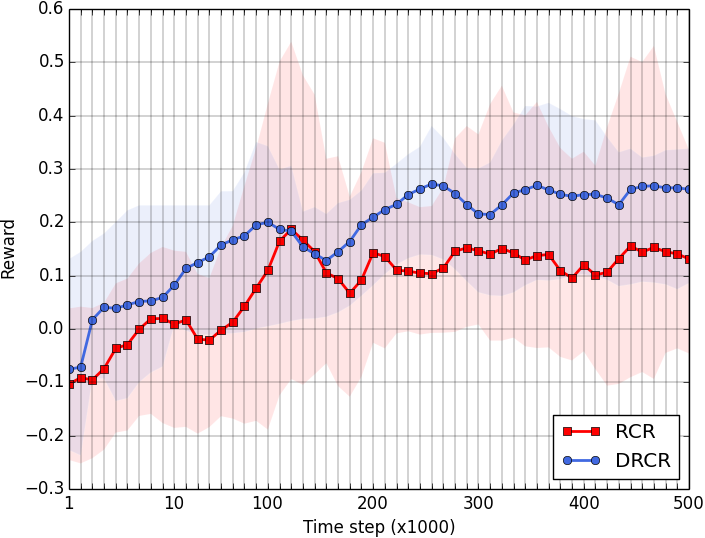}
    \end{minipage}
}
\subfigure[]{
    \begin{minipage}[b]{0.3\textwidth}
    \includegraphics[width=1\textwidth]{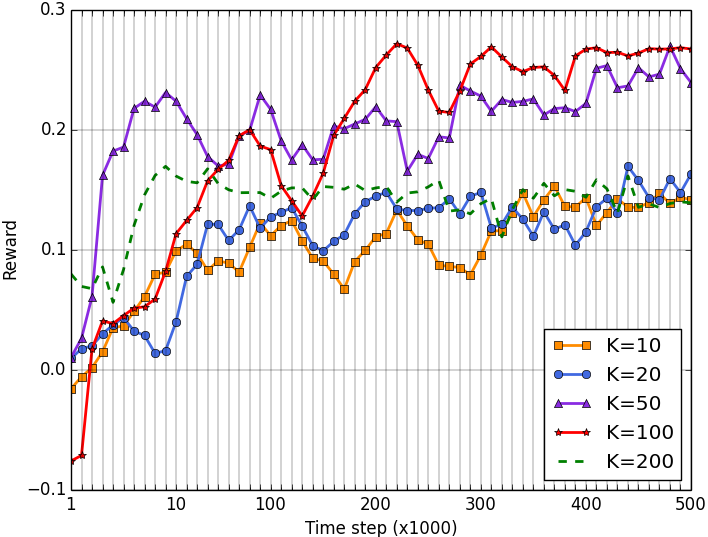}
    \end{minipage}
}
\subfigure[]{
    \begin{minipage}[b]{0.3\textwidth}
    \includegraphics[width=1\textwidth]{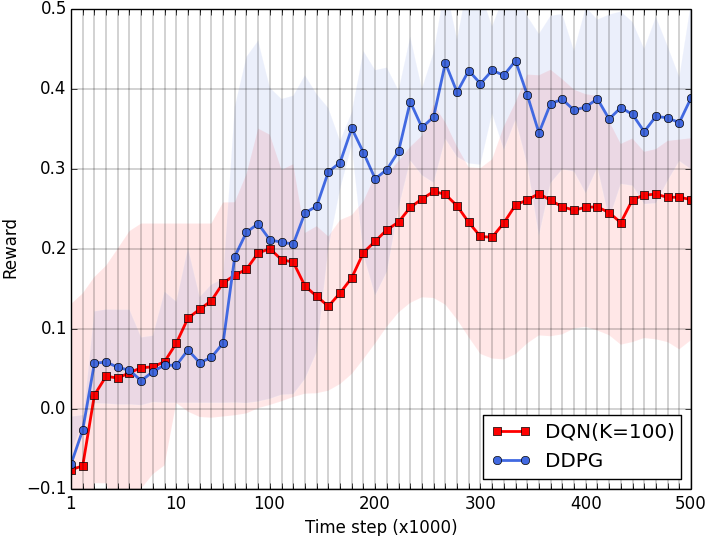}
    \end{minipage}
}
\caption{(a) Offline evaluation for different reward functions. (b) Offline evaluation for DQN with different $K$ values. $K$ stands for the number of output discrete actions for DQN models. (c) Offline evaluation for DQN and DDPG comparison.}
\label{fig:evaluation}
\end{figure}

For offline experiments, we carefully chose over 40,000 SKUs of FMCGs from hundreds of different categories. We used 60 days selling records to form about 2,400,000 tuples. The first 59 days' records were used for pre-training and the last day's for evaluation. We first evaluate two different reward functions we discussed above, revenue conversion rate (RCR) and the difference of revenue conversion rate (DRCR). We set $\alpha = 0.01$, $\tau = 1$ and gradually increased $\gamma$ from 0.5 to 0.99. We use DQN model with $K=100$ for this evaluation and the result is shown in Figure \ref{fig:evaluation}a. We could see that,the policy with DRCR as reward function performs better and steadier than using RCR. 

Then we evaluate DQN model with different parameter $K$. we set $K =10, 20, 50, 100$ and $200$. The result is shown in Figure \ref{fig:evaluation}b. We could see that, for our experiment setup, DQN with $K=100$ performs best. For DQN and DDPG comparison, we set $\epsilon=0.05$ and $K=100$. The result is shown in Figure \ref{fig:evaluation}c. DDPG performs better than DQN after pre-trained with a certain number of demonstrations.

\subsection{Online experiments}\label{para:online_experiments}

\begin{figure}[ht]
\centering
\includegraphics[width=0.4\textwidth]{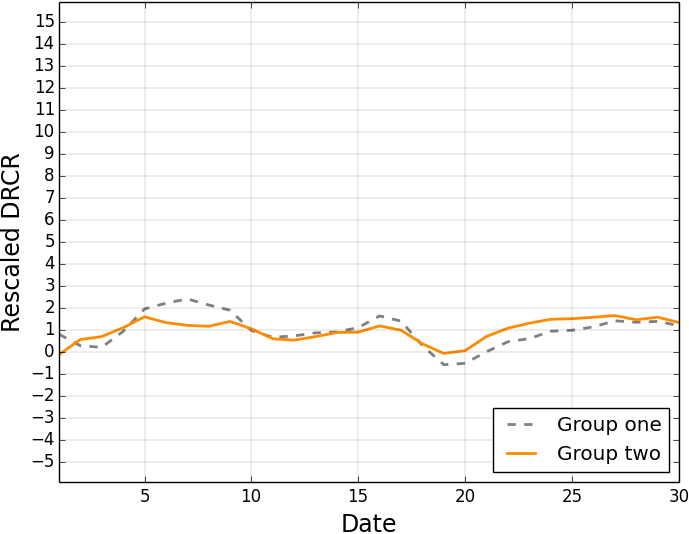}
\caption{ The DRCR for different groups of simi-products under same pricing policy within 30 days. The average of DRCR from group one is re-scaled to 1.00 and the average of group two becomes 0.99 after re-scaling.}
\label{fig:simi_product}
\end{figure}

To evaluate different pricing policies online, we need to eliminate the fluctuation of the market environment, causing by seasonality of demand or marketing operations. Since online A/B testing is not suitable for E-commerce platform as we discussed, we follow the idea of the difference-in-differences (DID) techniques (\cite{bertrand2004much},\cite{abadie2005semiparametric}), to evaluate the effects of different policies on relevant outcome. We first defined \textit{simi-products} on E-commerce platform for our experiment, the products with the same brand, same category and similar selling behaviours. We have found that, using the same pricing policy, two groups of simi-products could have very close DRCR with certain parameter $\tau$, even if two groups have different total revenues or revenue conversion rates. We set $\tau = 365$ here, making DRCR also represents the year-on-year growth of the revenue conversion rates, in order to eliminate the season fluctuation and influence from annual marketing strategies. Figure \ref{fig:simi_product} demonstrates that, two different groups of simi-products pricing by same policy have very close DRCR within 30 days. It is worth mentioning that, the DRCR with $\tau = 365$ here is an index for DID evaluation, rather than a reward function.

\textbf{Pricing for markdown season.} During the first part of online experiment, our agent prices 500 SKUs of luxury products (mainly handbags and clothes). Each product has around 10 items in stock and the aim of this markdown season is to maximize the total profit conversion rate. Therefore we use the profit conversion rate as our reward function, $r_{i,t} = profit_{i,t}/uv_{i,t}$. Another set-up in this online experiment is that, the agent is required to output a discrete value form 1 to 9 representing the discount rate from 10\% to 90\% in markdown season. So we applied discrete action model, DQN with $K=9$ here for these luxury products markdown pricing. We set $\tau = 1$ for reward function and $D=90$, $\alpha=0.01$ and $\gamma=0.99$ as pre-training parameters.

\begin{figure}[ht]  
\centering
\subfigure[]{
    \begin{minipage}[b]{0.45\textwidth}
    \includegraphics[width=1\textwidth]{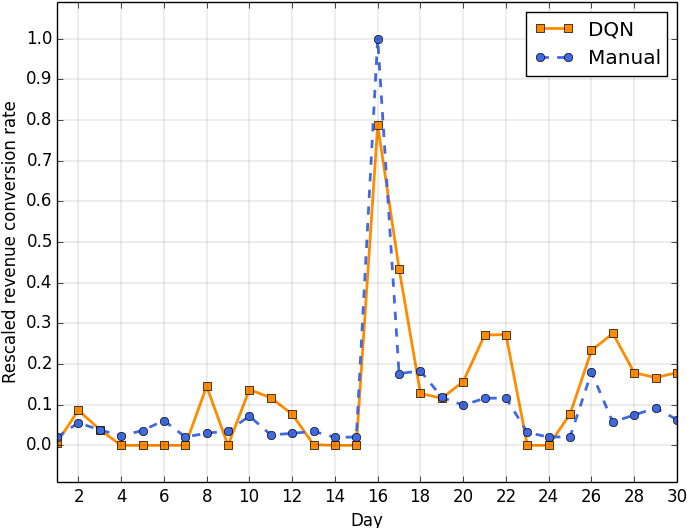}
    \end{minipage}
}
\subfigure[]{
    \begin{minipage}[b]{0.45\textwidth}
    \includegraphics[width=1\textwidth]{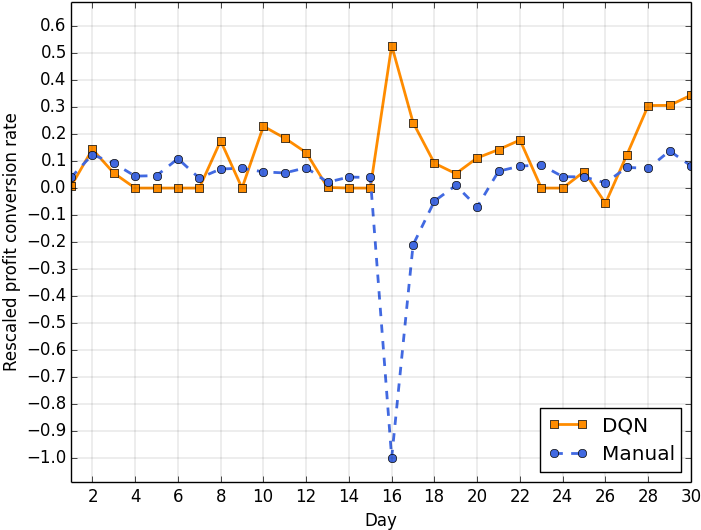}
    \end{minipage}
}
\caption{(a) Re-scaled revenue conversion rate from products priced by DQN and managers manually. From day 1 to day 15, DQN group and manual group both had a revenue conversion rate of 0.04. From day 16 to day 30, DQN group achieved a revenue conversion rate of 0.22 on average and manual group's average conversion rate is 0.16 for this period (with day 16 manual group's revenue conversion rate re-scaled to 1). (b) Re-scaled profit conversion rate from products priced by DQN and managers manually. From day 1 to day 15, the average profit conversion rate for both DQN group and manual group are 0.06. From day 16 to day 30, DQN group obtained an average profit conversion rate of 0.16, while manual group's average profit conversion rate dropped to -0.04 (with day 16 manual group's profit conversion rate re-scaled to -1)}
\label{fig:mlh}
\end{figure}

There is another group containing 2000 SKUs of simi-products pricing manually in the same period, which is regarded as the benchmark group for DID evaluation. The result of the experiment is demonstrated in Figure \ref{fig:mlh}a and Figure \ref{fig:mlh}b.

In the first 15 days of this online experiment, these products were in their daily prices. It shows that two groups of products performs alike, which confirms our investigation of the simi-product above. Then the markdown season started at the 16th day and lasted for 15 days. We could see that both groups boosted the revenue conversion rate at the beginning of the markdown season (day 16). Then in day 21 and 22 as well as day 25 to 30, DQN group successfully pull up the revenue conversion rate again beating manual group. Comparing with the profit conversion rate in Figure \ref{fig:mlh}b, it is more clear that, manual pricing method pull up the revenue by setting the prices lower than the cost, causing negative profit. DQN pricing policy successfully keeps positive profit at most of the markdown season. It is interesting that, at day 26, DQN also priced the products to negative profit rate, and got negative immediate reward. But this action pulled up both the revenue and profit conversion rate in the rest of the markdown season, achieving good total profit. 

\textbf{Pricing for daily sales.} To verify the effectiveness of reward function in Eq.(\ref{con:reward_function}), we set up another online experiment, to price supply unlimited FMCGs. The whole experiment contains two parts and each part lasted for 30 days. We chose 1000 SKUs of FMCGs from 200 different categories selling on Tmall.com as our experiment group (group two in Figure \ref{fig:simi_product}). As for DID evaluation, we matched 3000 SKUs of simi-products also selling on the platform as the control group (group one in Figure \ref{fig:simi_product}). The control group was pricing by some managers. Due to the huge amount of the products to be priced manually, their policy is usually set up a price at the beginning of the month, and then change some of them if their revenues are below their expectations. 

\begin{figure}[ht]  
\centering
\subfigure[]{
    \begin{minipage}[b]{0.45\textwidth}
    \includegraphics[width=1\textwidth]{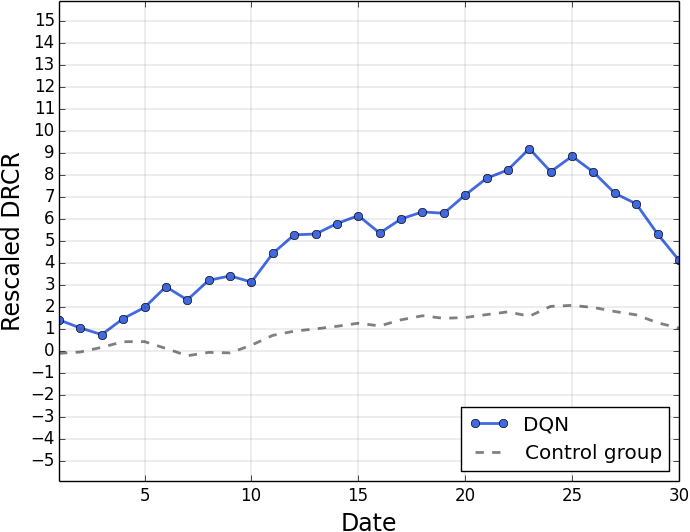}
    \end{minipage}
}
\subfigure[]{
    \begin{minipage}[b]{0.45\textwidth}
    \includegraphics[width=1\textwidth]{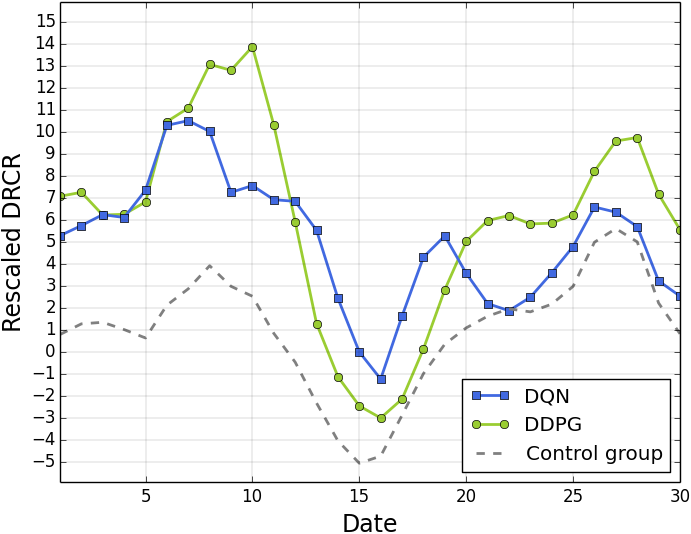}
    \end{minipage}
}
\caption{(a) The DRCR for DQN pricing group and manual pricing group within 30 days. The averages of DQN groups is 5.10, with the average of the control group re-scaled to 1.00. (b) Comparing the DRCR for DQN pricing group, DDPG pricing group and manual pricing group. The averages of DDPG and DQN pricing groups are 6.07 and 5.03 respectively with the average of the control group re-scaled to 1.00.}
\label{fig:tmcs}
\end{figure}

We set $K=100$ for DQN model and set $\alpha=0.01$ and $\gamma=0.99$. But as the FMCGs' behaviour changes more rapidly than the luxury products', we set $D = 30$ during pre-training, shorter than we set in markdown pricing experiment. In the first 30 days of experiment, we investigated the behaviour of DQN pricing policy by comparing it with the control group. DRCR within 30 days for two groups are shown in Figure \ref{fig:tmcs}a. We could see that, DQN group outperformed control group. 

Then we divided the experiment group randomly into two groups for testing DQN and DDPG, while kept the group one still as the control group. The result is shown in Figure \ref{fig:tmcs}b. During this part of experiment, we encountered some daily management activities (with some coupons given out). These activity only influenced less than $10\%$ of the total revenue. As the relationship between simi-products we mentioned above could still be observed, we did not stop the field experiment but kept observing the behavior of two models encountered the fluctuation of the environment. Both DRL methods outperformed control group while DDPG performed better.

\section{Conclusion and discussion}\label{conclusions}
In this work, we proposed a deep reinforcement learning framework for dynamic pricing on E-commerce platform. We defined the pricing process as a Markov Decision Process and then defined the state space, discrete and continuous action space, and different reward function for different pricing applications. We applied our methods for pricing policies and applied to online pricing in real time. We first apply deep reinforcement learning method for pricing products in a markdown season. The field experiment showed that it outperformed the manual markdown pricing strategy. As daily pricing for FMCGs, we design a systematic mechanism for online pricing policy evaluation, to address the legal issue of A/B testing for different pricing strategies. We showed that pricing policies from DDPG and DQN outperformed other pricing policies significantly.

This work is the first to use deep reinforcement learning for dynamic pricing problem on E-commerce platform, pricing thousands of SKUs of products in real-time. In this work, there are a few constraints could be removed. First, our pricing framework trains each product separately. As a result, the low-sales-volume products may not have sufficient training data. This could be solved by clustering similar products and using transfer learning to price the products in the same cluster. Meta-learning may also help for this problem. Second, our framework outputs pricing policy for each product separately. However, sometimes we hope to price different products together to form certain marketing strategies. This may be solved by a combinatorial action space. Third, in our pricing framework, we take only the features related to the products to describe the environment state. In the future, we would try to take more kinds of features into consideration for pricing under more specific scenarios, e.g., promotion pricing or membership pricing.

\section*{Acknowledgement}
We would like to thank Sentao Miao, Huiqiang Mao, Miaolan Xie, Yangsheng Ji and others at Alibaba Supply Chain Platform for their helpful comments.

\bibliographystyle{unsrt}  

\bibliography{references} 

\end{document}